\documentclass[10pt,conference]{IEEEtran}
\IEEEoverridecommandlockouts

\usepackage{cite}
\usepackage{amsmath,amssymb,amsfonts}
\usepackage{algorithmic}
\usepackage{graphicx}
\usepackage{textcomp}
\usepackage[english]{babel}
\usepackage{amsmath}
\usepackage{graphicx,epstopdf}
\usepackage[colorlinks=true, allcolors=blue]{hyperref}
\usepackage[english]{babel}
\usepackage[ruled,vlined]{algorithm2e}
\usepackage{amsmath}
\usepackage{booktabs}
\usepackage{hyperref}
\usepackage{graphicx}
\usepackage{amsmath,amssymb}
\usepackage{latexsym}
\usepackage{eurosym}
\usepackage{units}
\usepackage{mathtools}
\usepackage{lscape} 
\usepackage{amssymb}
\usepackage{xcolor}
\usepackage{lipsum}
\usepackage{optidef}
\usepackage{multirow}
\usepackage{booktabs}

\def\BibTeX{{\rm B\kern-.05em{\sc i\kern-.025em b}\kern-.08em
    T\kern-.1667em\lower.7ex\hbox{E}\kern-.125emX}}

\IEEEoverridecommandlockouts\IEEEpubid{\makebox[\columnwidth]{ 979-8-3503-1090-0/23/\$31.00~\copyright~2023 IEEE \hfill} \hspace{\columnsep}\makebox[\columnwidth]{ }}

\begin{document}

\title{

Multi-UAV Speed Control with Collision Avoidance and Handover-aware Cell Association: DRL with Action Branching
\thanks{The authors emails are \{zjyan, hinat\}@yorku.ca; \{wael.jaafar, bassant.selim\}@etsmtl.ca.}
}

\author{%
  \IEEEauthorblockN{%
    Zijiang~Yan\IEEEauthorrefmark{1}, 
    Wael~Jaafar\IEEEauthorrefmark{2},
    Bassant~Selim\IEEEauthorrefmark{2} and
    Hina~Tabassum\IEEEauthorrefmark{1}%
  }%
  \IEEEauthorblockA{\IEEEauthorrefmark{1} York University, ON, Canada, \IEEEauthorrefmark{2} École de Technologie Supérieure (ÉTS), University of Quebec, QC, Canada}%
}

\maketitle
\raggedbottom
\begin{abstract}
This paper develops a deep reinforcement learning solution to simultaneously optimize the multi-UAV cell-association decisions and their moving speed decisions on a given 3D aerial highway.  The objective is to improve both the transportation and communication performances, e.g., collisions, connectivity, and HOs. We cast this problem as a Markov decision process (MDP) where the UAVs' states are defined based on their speed and communication data rates. We have a 2D transportation-communication action space with decisions like UAV acceleration/deceleration, lane changes, and UAV-base station (BS) assignments for a given UAV’s state. To deal with the multi-dimensional action space, we propose a neural architecture having a shared decision module with multiple network branches, one for each action dimension. A linear increase of the number of network outputs with the number of degrees of freedom can be achieved by allowing a level of independence for each individual action dimension. To illustrate the approach, we develop Branching Dueling Q-Network (BDQ) and Branching Dueling Double Deep Q-Network (Dueling DDQN). Simulation results demonstrate the efficacy of the proposed approach, i.e., 18.32\% improvement compared to the existing benchmarks.
\end{abstract}

\begin{IEEEkeywords}
Unmanned aerial vehicles,  HOs,   Deep Reinforcement Learning, speed, cell-association.
\end{IEEEkeywords}

\section{Introduction}

Unmanned aerial vehicles (UAVs) are gaining popularity  across a broad range of applications due to their mobility, flexible deployment, gradually decreasing production costs, and line-of-sight (LOS) channels. 
\cite{yu2022deep}. 
A UAV can either require cellular connectivity for its own use (UAV-UEs) or provide cellular coverage as a base station (BS). Nevertheless, controlling UAVs that operate beyond visual line of sight (BVLoS) requires reliable command and control which is crucial for mission safety and security.

Existing research primarily focuses on optimizing cellular link availability and quality of service (QoS) using reinforcement learning (RL) algorithms with no considerations to multi-UAV aerial traffic flow and motion dynamics of UAVs.  In \cite{cherif2022cellular}, the authors proposed a RL algorithm that considers disconnectivity, HOs, and energy consumption for trajectory planning and cell association in cargo UAVs. However, the algorithm's actions only consider the direction of motion with no speed and lane considerations. In \cite{ chen2020efficient}, the authors present strategies based on deep learning to predict HOs in mmWave communications and optimize HO rates and radio link quality for known UAV trajectories. However, these works have not considered the motion dynamics factors, such as acceleration, deceleration, and lane changes on the aerial highway. Furthermore, the existing works in \cite{cherif2021disconnectivity} mostly considered $Q$-learning and its variants which can lead to sub-optimal policies and slower convergence. In terms of transportation, achieving high performance for multi-UAV traffic flow and collision avoidance is crucial. On the communication side, UAVs require: \textbf{(i)} high data rates and \textbf{(ii)} minimal HO losses. Increasing speed can increase traffic flow but results in frequent HOs, which can negatively impact the communications between UAVs and base stations (BSs). Very recently, this trade-off has been investigated in the context of autonomous vehicles \cite{10077729,10001396}.

 Nevertheless, previous studies have not considered speed optimization of multiple UAVs on an aerial highway in conjunction with cell association, while considering collision avoidance, lane changes, and HO-aware wireless data rates. 

In this paper, we develop a deep RL (DRL) solution with action branching architecture to jointly optimize cell-association and multi-UAV flying  policies on a 3D aerial highway such that \textbf{(i)}  aerial traffic flow can be maximized  with the collision avoidance, and \textbf{(ii)} HO-aware data rates can be maximized. Specifically, we first cast this problem as Markov decision process (MDP) where a UAV state is modeled based on its speed and data rates. Moreover, to deal with the 2D communication-transportation action space, we develop a DRL solution with an action branching architecture in which a shared module coordinates among multiple network branches. In our case, the module performs 2D decision-making related to UAV acceleration/deceleration, lane changes, and UAV-BS assignments. 
To illustrate the approach, we proposed Branching deep Q (BDQ) network and Branching double deep Q-network (BDDQN)-based UAV agents.
The proposed BDQN offers improved exploration-exploitation trade-off, and enhances robustness and stability compared to conventional DQN.

\section{System Model}

\begin{figure}[ht]
\includegraphics[width=1\linewidth]{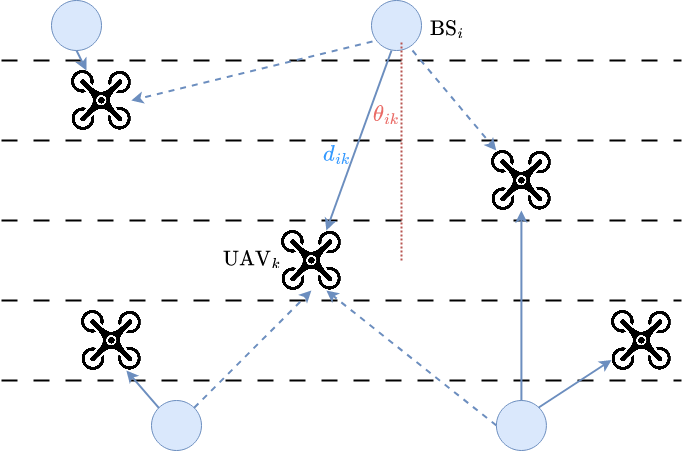}
\caption{Illustration of the proposed aerial network model (top view). Blue circles represent BSs; Solid/dash lines represent desired/interference link. 
}
\label{fig:BS_distributions_projection}
\end{figure}

As illustrated in Figs. \ref{fig:BS_distributions_projection}, 
we assume a 3D area where $N_U$ UAVs in a set $\mathcal{U}=\{ u_1, \ldots, u_{N_U}\}$ are flying along the defined 3D highway lanes, while being connected to terrestrial BSs. The latter are uniformly distributed on the targeted area and constitute a set $\mathcal{B}=\{b_1,\ldots, b_{N_R} \}$. To simulate the UAVs' movements on a given aerial highway, we consider the continuous intelligent driver model that models acceleration as in \cite{Treiber2013}. 
UAVs cannot fly above $h_{\max}=300$ m \cite{3gpp777}, and each UAV is identified by its location $\textbf{q}_{k}(t)=\left(x_k(t),y_k(t),h_k \right)$ at any time slot $t$, $\forall k \in \mathcal{U}$. Similarly, the BSs are defined by their locations $\textbf{q}_i=\left(x_i, y_i, h_i \right)$, $\forall i \in \mathcal{B}$. For the sake of simplicity, we assume that $h_i=0$ m, $\forall i \in \mathcal{B}$. The distance between BS $i$ and UAV $k$ is defined as $q_{ik}(t)=\sqrt{(x_k(t)-x_i)^2+(y_k(t)-y_i)^2+h_k^2}$ and the projected distance on the 2D plane (X,Y) is $d_{ik}(t)=\sqrt{(x_k(t)-x_i)^2+(y_k(t)-y_i)^2}$.  


\subsection{G2A Channel Model}

According to 3GPP \cite{3gpp777}, the ground-to-air (G2A) channel model is characterized by the BS's antenna gain, and the experienced path loss and line-of-sight (LoS) probability. 


\subsubsection{BS's antenna gain}
In cellular-connected aerial networks, UAVs rely on the radiating sidelobes to connect to terrestrial BSs. Hence, it is important to accurately model the 3D radiation pattern of BSs for cellular-connected UAVs. We opt here for the 3GPP antenna pattern model \cite{3gpp777} that mimics realistic antenna radiation patterns. Specifically, each BS is divided into three sectors, each equipped with cross-polarized antennas to create a uniform linear array (ULA). Each antenna element provides a gain up to $G_{\max}=8$ dBi through the direction of the main lobe \cite{3gpp777}. The antenna element pattern provides different gains on sidelobes depending on the azimuth and elevation angles of the associated UAV \cite{cherif2022cellular}. The latter are given by 
\begin{equation}
    G_{\mathrm{az}}(\phi_{ik}(t)) =\min \left\{ 12 \left( \frac{\phi_{ik}(t)}{\phi_\mathrm{3dB}} \right),\mathrm{G_m}  \right\},
\end{equation}
and
\begin{equation}
    G_{\mathrm{el}}(\theta_{ik}(t)) =\min \left\{ 12 \left( \frac{\theta_{ik}(t)}{\theta_\mathrm{3dB}} \right), \mathrm{SLA}  \right\},
\end{equation}
where $\phi_{ik}(t) = \arctan \left( \frac{h_k}{d_{ik}(t)}\right)$ and $\theta_{ik}(t) = \arctan \left(\frac{y_k(t)-y_i}{x_k(t)-x_i}\right)$ are the azimuth and elevation angles between BS $i$ and UAV $k$. $\phi_\mathrm{3dB}=\theta_\mathrm{3dB}=\frac{65\pi}{180}$ at 3dB bandwidths. In addition,  $\mathrm{G_m} $ and $\mathrm{SLA}$ are the antenna nulls thresholds, which are fixed at 30 dB in our study.   The antenna element gain is defined by \cite{cherif2022cellular}
\begin{flalign}
    G(\theta_{ik}(t), \phi_{ik}(t)) &= G_{\mathrm{max}} \\ 
    &- \min \{- (G_{\mathrm{az}}(\phi_{ik}(t)) + G_{\mathrm{el}}(\theta_{ik}(t)) ) , G_m \} \nonumber.
\end{flalign}
Assuming that BS $i$ has $N$ antennas inter-separated by half of the wavelength distance \cite{cherif2022cellular}, the array factor, denoted AF, of the ULA of BS $i$ towards UAV $k$ is expressed by
\begin{equation}
    \mathrm{AF}(\theta_{ik}(t)) = \frac{\sin(\frac{N\pi}{2} (\sin \theta_{ik}(t) - \sin \theta_i^d))}{\sqrt{N} \sin(\frac{\pi}{2} (\sin\theta_{ik}(t) - \sin\theta_i^d))},
\end{equation}
where $\theta_i^d$ is the down-tilt of BS $i$'s ULA. Finally, the array radiation pattern from BS $i$ towards UAV $k$ is written as
\begin{equation}
G_{ik}(t) = G(\theta_{ik}(t),\phi_{ik}(t)) + \mathrm{AF}(\theta_{ik}(t)), \; \forall i \in \mathcal{B}, \forall k \in \mathcal{U}.
\end{equation}

\subsubsection{LoS probability}
The likelihood of UAV $k$ having a LoS with BS $i$ primarily relies on the altitude of the UAV and the surrounding  environment. 
Assuming that $h_k \in [22.5,100]$ m, the probability of LoS is given by \cite{cherif2022cellular} is 
\begin{equation}
\small
    P_{\mathrm{LoS}}(q_{ik}(t))= 
\begin{cases}
1,\quad &d_{ik}(t) \leq d_1 \\
\frac{d_1}{d_{ik}(t)} + e^{-\frac{d_{ik}(t)}{p_1}}\left(1- \frac{d_1}{d_{ik}(t)} \right),\quad &\text{otherwise,}
\end{cases} 
\end{equation}
where $d_1 = \max\{460\log_{10}(h_k)-700,18\}$ and $p_1 = 4300\log_{10}(h_k)-3800$. If $h_k \in [100,300]$ m, $P_{\mathrm{LoS}}(q_{ik}(t)) = 1$. Thus, the probability of Non-LoS (NLoS) is written as $P_{\mathrm{NLoS}}(q_{ik}(t)) = 1 - P_{\mathrm{LoS}}(q_{ik}(t))$.




\subsubsection{Path loss}
For the sake of simplicity, we consider the mean path loss  since we focus here on the long-term operation of cellular-connected UAVs rather than the short term \cite{alzenad20173}. The probabilistic mean path loss between BS $i$ and UAV $k$ at time slot $t$ can be expressed by 
\begin{flalign}
    L_{ik}(t) &= L_i^{\mathrm{LoS}}P_{\mathrm{LoS}}(r_{ik}(t))  \\ &+L_i^{\mathrm{NLoS}}P_{\mathrm{NLoS}}(r_{ik}(t)), \forall i \in \mathcal{B},\nonumber
\end{flalign}
where  $L_i^{\mathrm{LoS}}$ and $L_i^{\mathrm{NLoS}}$ are the path loss related to LoS and NLoS communication links, respectively, as defined in \cite[Tables B-1 and B-2]{3gpp777}.


\subsection{Received Power and Achievable Data Rate Analysis}
Assuming that UAV $k$ has an omni-directional antenna, and using the G2A channel model, the average power received from BS $i$ can be expressed by
\begin{equation}
    P_{ik}(t) = P_T + G_{ik}(t) - L_{ik}(t) - P_n , \forall i \in \mathcal{B}, \forall k \in \mathcal{U},
\end{equation}
where $P_T$ is the transmit power of any BS $i$ and $P_n$ is the noise power (in dBm).
The quality of the link between UAV $k$ and BS $i$ is determined by the strength of the received signal from the latter, evaluated with $P_{ik}(t)$. However, since the aerial highways can be served by several terrestrial BSs with the same frequency, mainly due to the strong LoS between BSs and UAVs, then significant interference can be generated. Consequently, the quality of a communication link is rather evaluated using the signal-to-interference-ratio (SIR)\footnote{In practice, the quality of the link should be evaluated using the signal-to-interference-plus-noise-ratio (SINR). However, due to the significant interference generated in the considered system model, we ignore the noise's effect.}. The latter is written by
\begin{equation}
    \text{SIR}_{ik}(t) = \frac{P_{ik}(t)}{\sum_{\substack{j=1,\\ j \neq i}}^{N_R}P_{jk}(t)}, \forall i = 1 , \dots ,  N_R, \forall k=1, \dots,N_U.
\end{equation}
Assuming that all BSs use the same bandwidth $W_R$, then the achievable data rate between BS $i$ and UAV $k$ is given by 
 \begin{equation}
     R_{ik}(t) = W_R\log_2(1+\text{SIR}_{ik}(t)), \forall i \in \mathcal{B}, \forall k \in \mathcal{U}.
 \end{equation}

\subsection{Handovers}
Since a flying UAV $k$ can evaluate $P_{ik}(t)$ from neighbouring BSs, i.e., $i \in \mathcal{B}_{k}(t)$ where $\mathcal{B}_{k}(t)$ is the set of the closest $n_{\rm rf}$ BSs to UAV $k$ that can serve UAV $k$, i.e., BS $i \in \mathcal{B}_{k}(t)$ if $q_{ik}(t)\leq d_{th}$ and SIR$_{ik}(t)\geq \gamma_{th}$, where $d_{th}$ is the maximal communication distance between any BS and UAV $k$ and $\gamma_{th}$ is the UAV's reception sensitivity. Subsequently, a UAV can trigger a HO event whenever required, e.g., when SIR$_{i_0k}(t)< \gamma_{th}$, where BS $i_0$ is the one that UAV $k$ is currently associated with.    
To reflect HO events, let $c_k(t)$ be the index of the BS to which UAV $k$ is associated at time slot $t$ and $\eta_k$ be the HO binary variable, such that $\eta_k(t,t+1)=1$ if $c_{k}(t)\neq c_k(t+1)$ and $\eta_k(t,t+1)=0$ otherwise.  
Frequent HOs can severely impact the received SIR due to HO overhead and risk of HO ping-pong effect \cite{chen2020efficient}.




\section{Problem Formulation as MDP and Proposed DRL with Action Branching}
Given the described system model, we aim to collaborative optimize the autonomous motion of multiple UAVs travelling along a 3D highway, such that both the transportation and communication performances, e.g., collisions, connectivity, and HOs, are improved. First, we specify the state-action space and rewards of our system. Then, we present the proposed collaborative RL-based solutions to control the UAVs. These solutions are based on the BDQN and BDDQN algorithms.



\subsection{Observation and State Space}
The observation space $\mathcal{O}$ 
provides RL agents with the necessary information to take actions that result in rewards. For our system, our observation space is composed of transportation and communication observations.  The transportation observation space is known as kinematics and is included in the \textit{highway-env}  environment \cite{highway-env}. The kinematics observation consists of a $V \times F$ array that describes a list of nearby UAVs $V < N_U$ based on $F$ specific features, namely ($\textbf{q}_k(t), \textbf{v}_k(t), n_\mathrm{rf}$), where $\textbf{v}_k(t)=[v_k^x(t),v_k^y(t)]$ is the directional speed of UAV $k$ on (X,Y) plane at time $t$, $v_k^x(t)$ represents the longitudinal speed, and $v_k^y(t)$ represents the latitudinal speed. $n_\mathrm{rf}$ is number of feasible BSs in radius of 1000m of target UAV. The feature values may be normalized within a predetermined range, with the normalization relative to the UAV that is about to take an action. 
Moreover, each UAV $k$ observes communications-related features such as the received SIR levels from BSs in $\mathcal{B}_k(t)$.  
The presence of several UAVs allows to simulate different traffic flow scenarios as in \cite{10077729}.
Indeed, as the density of UAVs in the highway increases, higher competition is expected to connect to the best terrestrial BSs among them. This holds true assuming that BS $i$ cannot be associated to more than $Q_i$ UAVs at the same time, $\forall i \in \mathcal{B}$. Thus, each BS $i$ has to continuously keep track of the number of associated UAVs to it all the time, denoted $n_i(t)$, $\forall i \in \mathcal{B}$.
Consequently, a state $s_t$ for an RL agent at UAV $k$ is constituted from several observations as $s_t=( \textbf{q}_1(t), \ldots, \textbf{q}_V(t), \textbf{v}_1(t), \ldots, \textbf{v}_V(t), n_\mathrm{rf}, \text{SIR}_{1k}(t),$ $\ldots,\text{SIR}_{n_{\mathrm{rf}}k},Q_1, \ldots, Q_{n_{\mathrm{rf}}},n_1(t),\ldots,Q_{n_{\mathrm{rf}}}(t))$.  



\subsection{Action Space}
\label{sec:2daction}
At each time step $t$, UAV $u_k$ selects action $a_t=(a_t^{\rm tran}, a_t^{tele}) \in \mathcal{A}_{\rm tran} \times \mathcal{A}_{\rm tele}$, where $a_t^{\rm tran}$ is the moving transportation action, i.e., trajectory action and $a_t^{tele}$ is the communication-related action, i.e., association with a terrestrial BS.
$\mathcal{A}_{\rm tran}=\{a_{\rm tran}^1,\ldots,a_{\rm tran}^5\}$, where $a_{\rm tran}^1$ is the change lane to the left lane action, $a_{\rm tran}^2$ is maintaining the same lane, $a_{\rm tran}^3$ is the change lane to the right one, $a_{\rm tran}^4$ is accelerating within the same lane, and $a_{\rm tran}^5$ is decelerating within the same lane. Similarly, the communication action space for $u_k$ at time $t$ can be given by $\mathcal{A}_{k,\rm tele}(t)=\left\{ c_k^1(t), \ldots, c_k^{n}(t) \right\}$, where $c_k^i(t)$ is the $i^{th}$ potential BS to be associated with.
Based on the quota of each BS $Q_i$, UAV computes a \textit{weighted rate metric}, denoted WR, that encourages traffic load balancing between BSs and discourages unnecessary HOs. It is expressed by
\begin{equation}
    \text{WR}_{ik}(t) = \frac{R_{ik}(t)}{\min \left(Q_i, n_i(t) \right)} (1 - \mu), \forall i=1,\ldots,n,
\end{equation}
   where $\mu$ denotes the HO penalty, written as
    \begin{equation}
    {
    \mu = 
    \begin{dcases}
    0.1, & \text{if HO is triggered,} \\
    0, & \text{otherwise.} 
    \end{dcases}
    }   
    \end{equation}
This criterion will be taken into account to further reduce the set of potential BSs. Specifically, the 
final BSs' set should be composed with $n \leq n_{\rm rf}$ candidates, which belong to $\mathcal{B}_k(t)$, satisfy   
$n_i(t)<Q_i$, and obtained the best WSIR values. 

The UAV computes $P_{{ik}}$ by substituting $\mu=0$ and chooses to connect to BS with the maximum  ${T}_{{ik}}$, if $Q_j \geq n_i(t)$, Otherwise, the UAV recursively selects the next vacant best-performing BS. 

\subsection{Reward Function Design}
The definition of the associated reward function is directly related to the optimization of both UAV transportation and communication performances. 

\subsubsection{UAV Transportation Reward}
We define the UAV transportation reward as follows \cite{highway-env}:
\begin{equation}
    r^{\mathrm{tran}}_{k} (t)=  \omega_1 \left( \frac{||\mathbf{v}_k(t)|| -v_\mathrm{min}}{v_\mathrm{max}-v_\mathrm{min}} \right)- \omega_2 \cdot \delta, \forall k \in \mathcal{U},
\end{equation}
where $v_{\min}$ and $v_{\max}$ are the minimum and maximum speed limits, and $\delta$ is the collision indicator. $\omega_1$ and $\omega_2=1-\omega_1$ are the weights that adjust the value of the UAV transportation reward with its collision penalty.
It is important to note that negative rewards are not allowed since they might encourage the agent to prioritize ending an episode early, by causing a collision, instead of taking the risk of receiving a negative return if no satisfactory trajectory is available.

\subsubsection{UAV Communication Reward}


We define the communication reward as follows:
\begin{equation}
    r^{\mathrm{tele}}_{k}(t)= \omega_3 R_{i_0k}(t) \left(1- \text{min}(1,\xi_k (t))\right),
\end{equation}
where  $R_{i_0k}(t)$ is the achievable data rate when associated with BS $i_0$, and  $\xi_k(t)$ is the HO probability, computed by dividing the number of HOs accounted until the current time $t$ by the time duration of previous time slots in the episode. 

\begin{figure*}[t]
\label{fig:dqn_comparison}
\centering
\begin{tabular}{lccccc}
\includegraphics[width=0.7\linewidth]{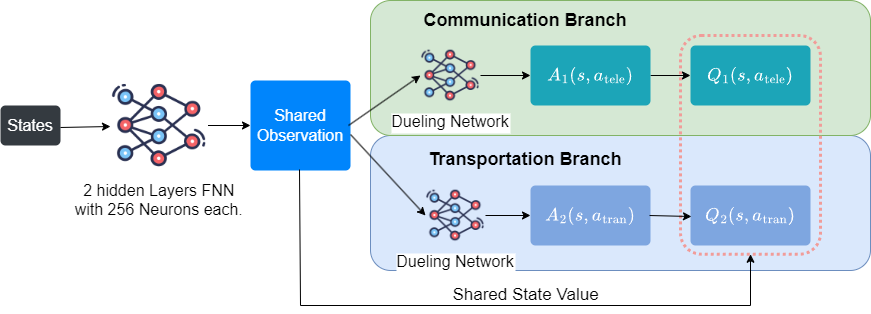}\hspace{-1cm}&\hspace{0.25cm}
\end{tabular}
\vspace{-2mm}
\caption {Proposed action
branching architecture. Shared module 
computes a latent representation of the input state and
passes it forward to action branches. } 
\label{fig:bdq_agent}
\end{figure*}

\subsection{Proposed Branching Dueling Q-Network-based Methods}
The use of discrete-action algorithms has contributed to many recent successes in deep reinforcement learning. However, implementing these algorithms in high-dimensional action tasks is challenging due to the exponential increase in the size of action space. In our study, for each time step $t$, we need to apply both communication action and transportation action on $N_U$ RL agents. 
To cope with such complex action design, authors of \cite{tavakoli2018action} introduced a novel RL agent based on branching dueling Q-network (BDQ) and illustrate the performance of branching deep Q-network (BDQN) or dueling double deep Q-network (BDDQN). BDQ features a decision module shared among multiple network branches, each corresponding to an action dimension, e.g., the transportation and communication action dimensions in our work. This approach allows for independent handling of each individual action dimension, resulting in a linear increase in the number of network outputs with the degrees of freedom. It also demonstrates the importance of the shared decision module in coordinating the distributed action branches. In this work, we take advantage of this method by deploying BDQ agents at the UAVs, and each of them makes actions branching for $\mathcal{A}_{\rm tran}$ and $\mathcal{A}_{\rm tele}$. 

\begin{algorithm}[ht!]
\caption{Proposed BDQN/BDDQN for Multi-UAV Speed Control and BS Association}
\SetAlgoLined
\KwResult{Action function $Q_\theta$ and Policy $\mathcal{J}$ }
\KwData{$Q$-network, Experience replay memory $\mathcal{D}$, mini batch-size $m$}
\textbf{Initialization:} $\mathcal{D} \gets \mathbf{0}$,  Q-network weights $\theta \gets \mathbf{0}$, Target network $\theta^* \gets \theta$, $Q(s,a)$, UAVs, BSs \\
\While{$\mathrm{episode} < \mathrm{episode \ limit \ or \ run \ time}< \mathrm{time \ limit}$}{
  $t \gets 0$,
  $s_t \gets \mathrm{horizon \ limit} $\\
\While{$t \leq \mathrm{horizon \ limit} $}{Each UAV $k$ selects $a_t$ by $\epsilon$-greedy search as  $a_t$:
\begin{align*}
\begin{cases} 
\text{Select $a_t$ from } \mathcal{A} & \text{prob. } \epsilon \\
\text{Select } a_t=\max_{a \in \mathcal{A}}{Q_{\theta}(s_t,a_t)} & \text{prob. } (1 - \epsilon)
\end{cases}
\end{align*}
Extract $a^{\mathrm{tran} }_{t}$ and $a^{\mathrm{tele} }_{t}$  from $a_t$ and apply them to UAV $k$;\\
\textbf{State-value Estimator:} Apply Eq. \ref{eqn:common_state_value_estimator} to compute $Q_d(s,a_d)$ \\
Store $(s_t,a_t,s_{t+1},r^{\mathrm{tran}}(t),r^{\mathrm{tele}}(t))$ to $\mathcal{D}$;\\
\textbf{Experience Replay:} sample transitions mini-batch in $\mathcal{D}$ $(s_k,a_k,r_k,s'_k)$ where  $k \in m$;\\
\textbf{Set target-$Q$ function:} Set $\hat{y}_k$ by Eq. \ref{eqn:target_function} 

\textbf{Set real $Q$-function:} $ y_k = {Q}(s_t,a_t;\theta)$;\\
Compute loss: $\mathcal{L}(\theta) =$ \\
$  \mathbb{E}_{(s_t,a_t,s_{t+1},R_t) \sim \mathcal{D}} \left[ \frac{1}{m} \sum_{k \in m} (y_k - \hat{y}_k)^2 \right]$;
\\
Perform gradient descent step by minimizing loss $\mathcal{L}$;
$\theta \gets \theta - a_{t} \cdot \mathcal{L}(\theta) \cdot \triangledown_{\theta}{y_k}$;\\

Update the deep-$Q$-network weights $\theta \gets \theta^*$
  }
   Policy $\mathcal{J}$ updated in terms of $Q$
 }
\end{algorithm}



According to \ref{sec:2daction}, we have two action dimensions and a total of $5 \times n$ sub-actions for each UAV at each time step. For an action dimension $d \in \{1,2\}$, each individual branch Q-value on state $s \in S $ and sub-action $a_d \in \mathcal{A}_d$ ($\mathcal{A}_1=\mathcal{A}_{\rm tran}$ and $\mathcal{A}_2=\mathcal{A}_{\rm tele}$) is defined by 
\begin{equation}
\label{eqn:common_state_value_estimator}
 Q_d(s,a_d) = V(s) + (A_d(s,a_d) - \max_{a'_d \in A_d}A_d(s,a'_d)), \forall d \in \{1,2\}. 
\end{equation}
Each sub-action affects the aggregating layer of $Q_d$ regard to dimension $d$. Based on the double DQN algorithm, we update the state-value estimator and loss function as in  \cite{tavakoli2018action}. 
We also adopt  the common state-value estimator based on dueling architecture. Dueling architecture reduces similar action redundancies and learning is shared by two branches visualized in Fig. \ref{fig:bdq_agent}. The target $Q$ function  $\hat{y}_k$ in BDDQN defined by \footnote{For the sake of simplicity, we define $\hat{y}_k = r^{\mathrm{tele}}_{k}(t) +  r^{\mathrm{tran}}_{k}(t) +  \frac{\gamma}{2} \sum_d Q_d^-\operatorname*{argmax}_{a_d' \in \mathrm{A}_d} (Q_d(s_k,a_d(t)))$ for BDQN. BDDQN depends on both Q-network and the target network while BDQN only rely on $Q$-network. $Q$-network, target network are designed for agent action selection and agent action evaluation respectively. }
\begin{equation}
\label{eqn:target_function}
   \hat{y}_k = r^{\mathrm{tele}}_{k}(t) +  r^{\mathrm{tran}}_{k}(t) +  \frac{\gamma}{2} \sum_d Q_d^-(s_k',\operatorname*{argmax}_{a_d' \in \mathrm{A}_d} (Q_d(s_k,a_d(t))) 
\end{equation}
where $Q_d^-$ is the branch $d$ of the target network $Q^-$.
The operation of the proposed BDQN/BDDQN-based approaches are summarized within Algorithm 1. DQN aims at compute weight sum $Q$ values for each aggregate actions tupple. This approach is eager to contribute unbalanced trade off between transportation reward and communication reward. However, The benefit for BDQN is finding optimized $Q_d$ in terms of $d \in \mathrm{A}_d$, which draws 2 optimal policies regarding to communication and transportation perspectives.


 \begin{figure*}[ht]
\centering
\begin{tabular}{lccccc}
\includegraphics[width=0.3\linewidth]{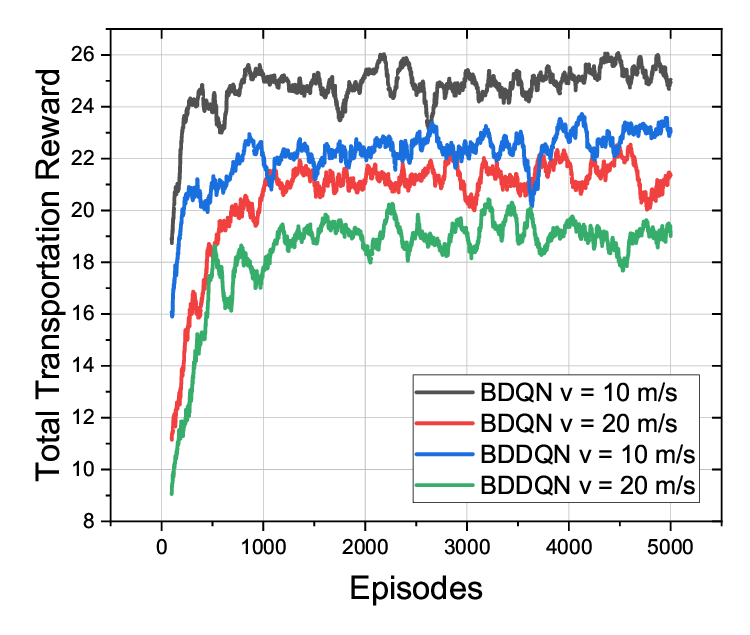}\hspace{-1cm}&\hspace{0.25cm}

\includegraphics[width=0.3\linewidth]{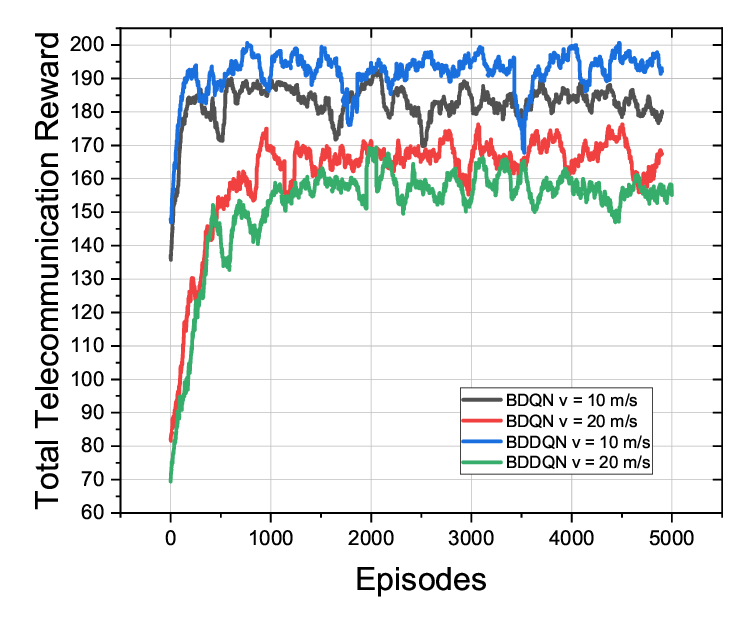}\hspace{-1cm}&\hspace{0.25cm}
\includegraphics[width=0.3\linewidth]{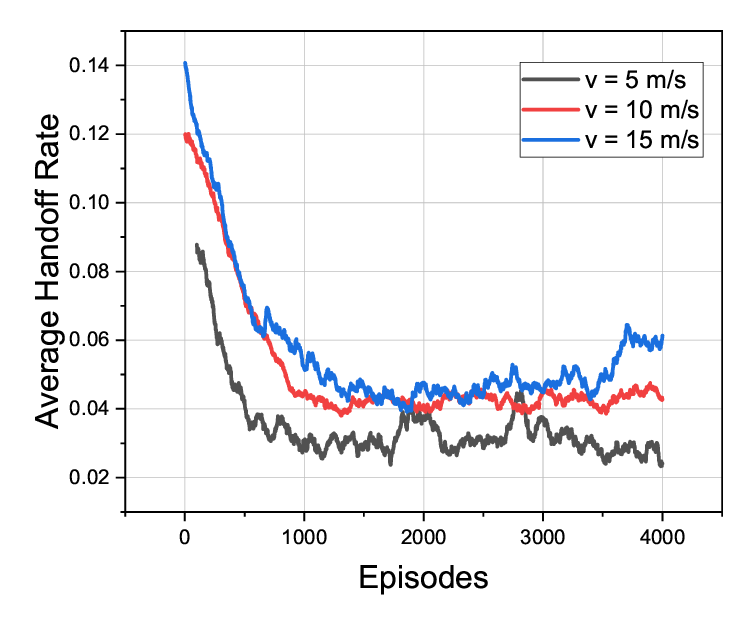}&\\
\qquad \qquad\qquad \qquad (a)  & (b) & (c) 
\end{tabular}
\vspace{-2mm}
\caption {UAVs performances ($15$ BSs, different $v$): (a) Total transportation reward (b) Total communication reward (c) HO rate (BDDQN). }
\label{fig:episodes-rewards}
\end{figure*}

\begin{figure*}[ht]
\centering
\begin{tabular}{lccccc}
\includegraphics[width=0.3\linewidth]{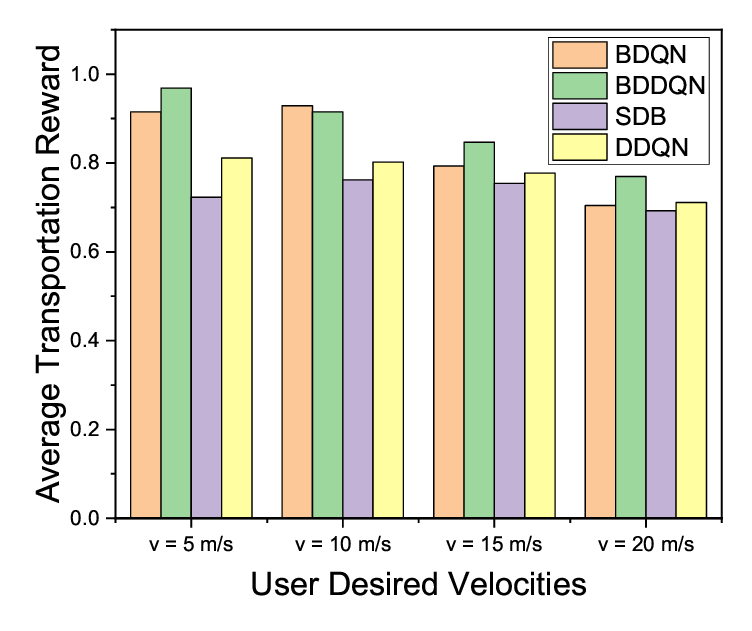}\hspace{-1cm}&\hspace{0.25cm}
\includegraphics[width=0.3\linewidth]{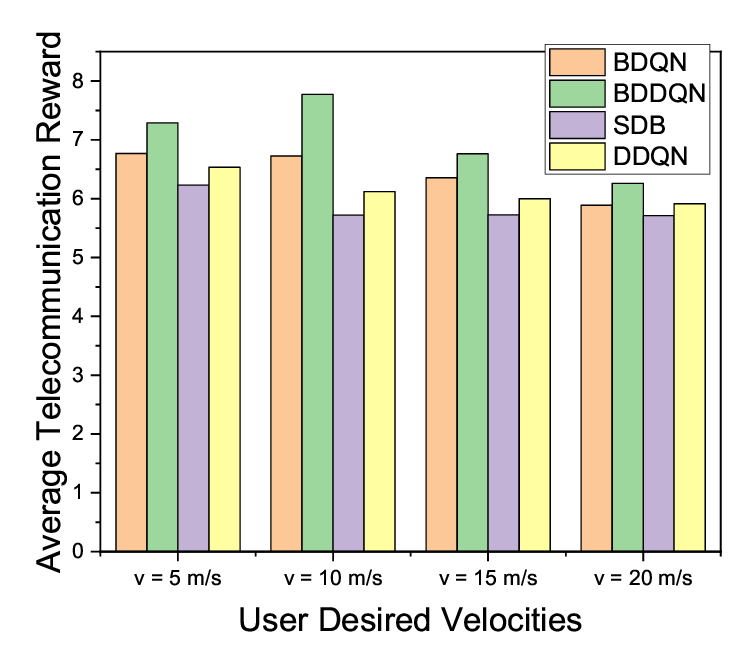}\hspace{-1cm}&\hspace{0.25cm}
\includegraphics[width=0.3\linewidth]{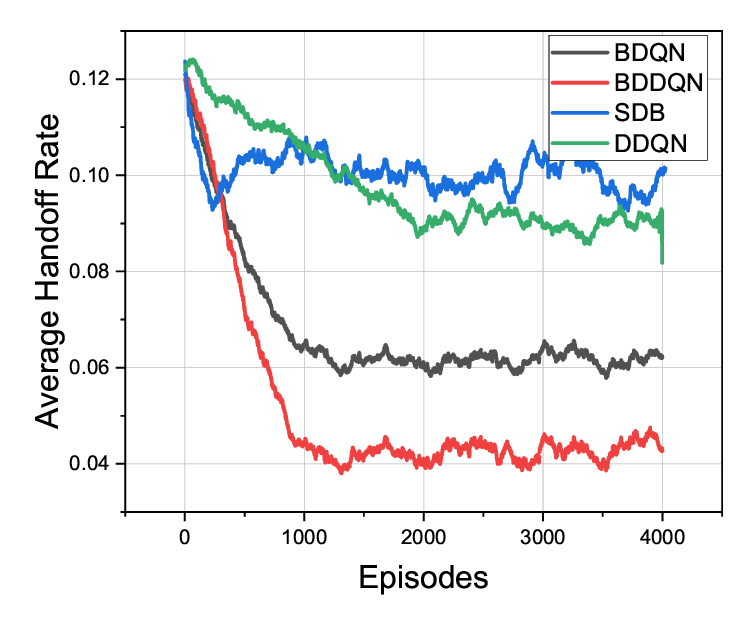}\hspace{-1cm}&\hspace{0.25cm}
&\\
\qquad \qquad\qquad \qquad (a)  & (b) & (c) 
\end{tabular}
\vspace{-2mm}
\caption{UAVs performances ($15$ BSs, $v=10$ m/s): (a) Avg. transportation reward (b) Avg. communication reward (c) Avg. HO rate. }
\label{fig:velocities-rewards}
\end{figure*}

 \begin{figure*}[ht]
\centering
\begin{tabular}{lccccc}
\includegraphics[width=0.3\linewidth]{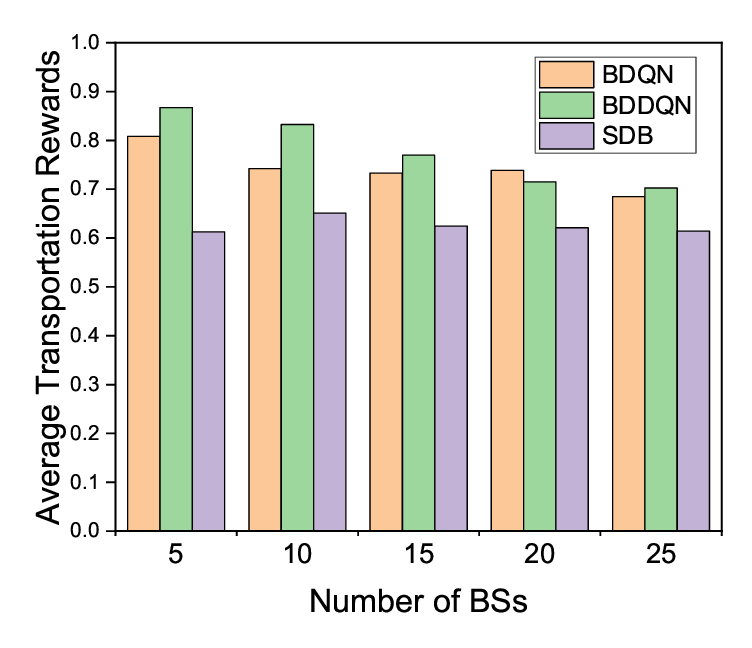}\hspace{-1cm}&\hspace{0.25cm}
\includegraphics[width=0.3\linewidth]{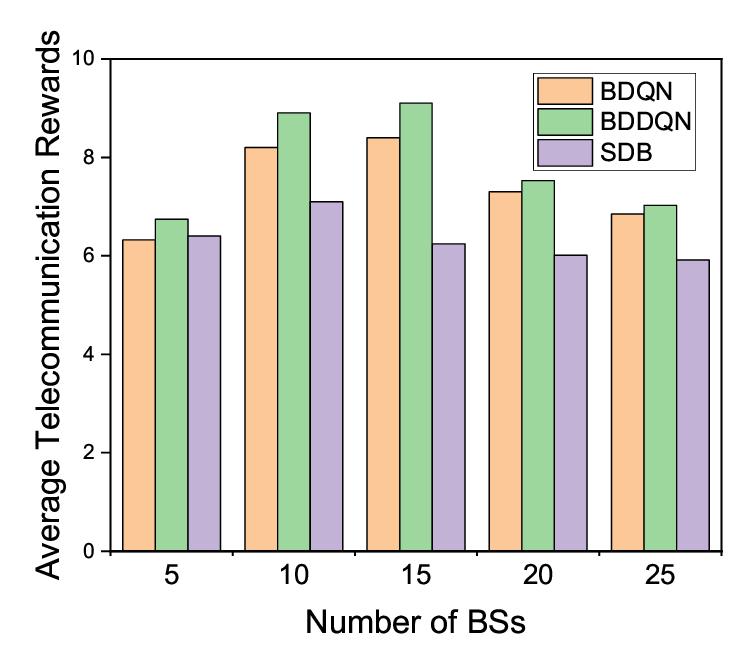}\hspace{-1cm}&\hspace{0.25cm}
\includegraphics[width=0.3\linewidth]{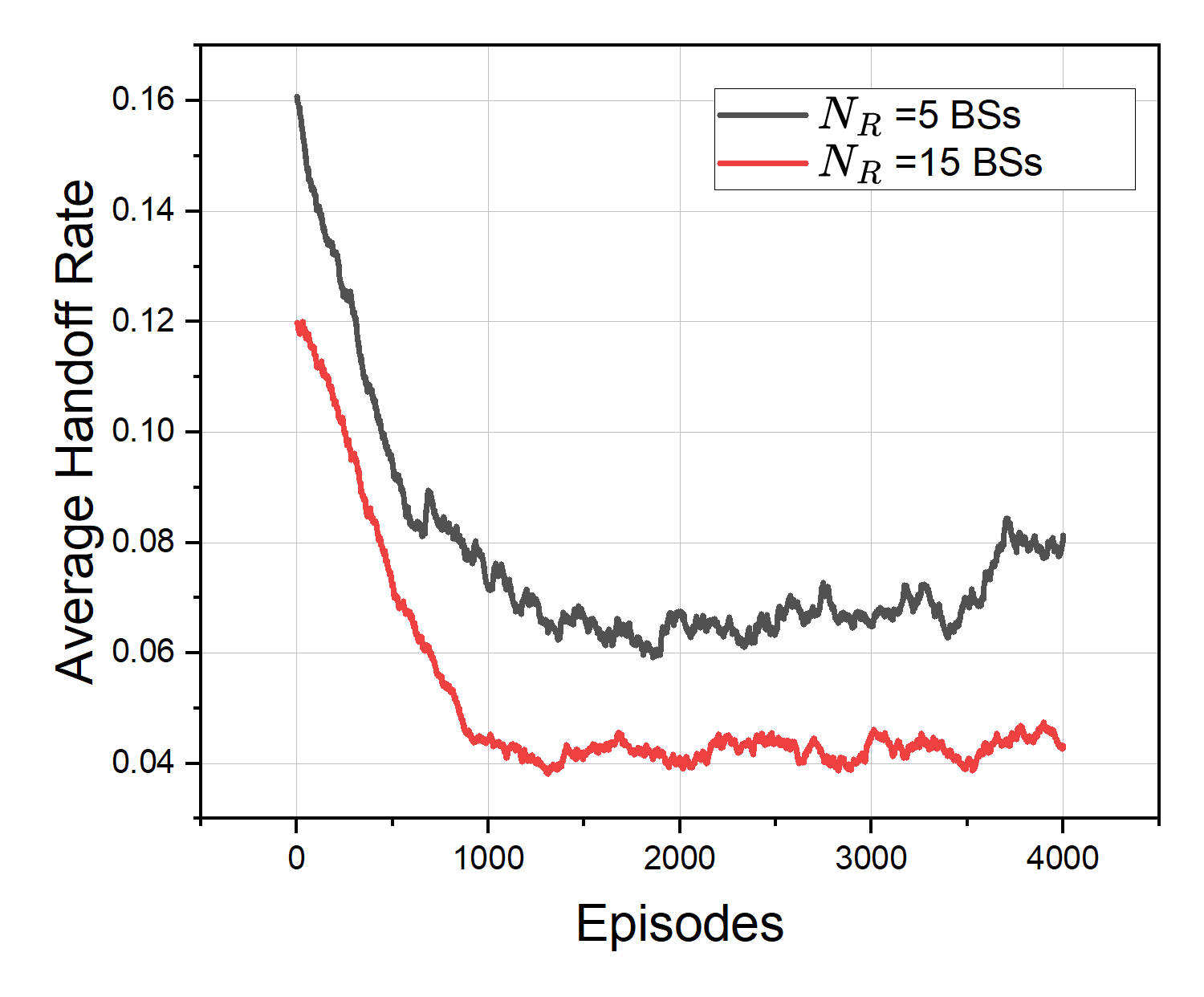}&\\
\qquad \qquad\qquad \qquad (a)  & (b) & (c) 
\end{tabular}
\vspace{-2mm}
\caption {UAVs performances ($v=10$ m/s): (a) Avg. transportation reward (b) Avg. communication reward (c) Avg. HO rate (BDDQN). }
\label{fig:nrbss-rewards}
\end{figure*}

\section{Numerical Results and Discussions}
In this section, we present the results of the suggested algorithms (BDQN and BDDQN) and emphasize the intricate relationships among wireless connectivity, handover rates, traffic flow of group UAVs, and the speed of UAVs.
Unless explicitly mentioned, we employ the subsequent simulation parameters. BSs operating on 2.1 GHz and maximum support $Q_R =5$ UAV users. We define $\eta_{\mathrm{LoS}}$ and $\eta_{\mathrm{NLoS}}$ as 1 and 20, respectively. There are 5 aerial highway lanes, where  $N_U=5$ UAVs fly at speeds between $v_{\min}=5$ m/s and $v_{\max}=20$ m/s. BS's transmission power $P_T$ is 40 dBm. The BDQN training learning rate $\alpha$, the discount factor $\gamma$, and the batch size are $5\times{10}^{-4},0.8$ and $32$, respectively.

The BDQ agent is represented in Fig. \ref{fig:bdq_agent}. To improve the performance of training and reduce the training complexity, we deploy a fully-connected feed-forward neural network (FNN) $N(s)$ with weights $\{\theta\}$ to approximate the $Q$-value for a given action and state \cite{10001396}. FNN takes the state as an input and outputs shared-observation in Fig. \ref{fig:bdq_agent}.
Since Q-values are real, the FNN performs a multivariate linear regression task. We apply ReLU activation function, i.e., $f_r(x) = \mathrm{max}(0,x)$, as the first layer. There are $2$ FNN hidden layers and $256$ neurons on each layer. There are single layers with $128$ neurons on each Branching dueling network on the branching stage. Linear activation function is at the output layer.

Fig. \ref{fig:episodes-rewards} states the training for UAV transportation rewards, communication rewards, and HO rate in terms of algorithms and target speed of UAVs, $v$. Fig. \ref{fig:episodes-rewards}(a) states transportation rewards reduce with higher UAV target speed. This is because higher speed contributes to higher collision occurrences. According to Fig. \ref{fig:episodes-rewards}(b), the communication rewards of BDQN and BDDQN are closer to each other for a given target speed while showing that a target speed of $10$ m/s is preferable over the $20$ m/s one. In any case, the gaps in communication rewards are small suggesting that the proposed algorithms tend to maximize the total communication reward regardless of the speed's impact. Finally, Fig. \ref{fig:episodes-rewards}(c) illustrates the HO rate convergence for BDDQN for different $v$. Clearly, after about 1000 episodes, BDDQN converges to HO values below 1\% since UAVs tend to avoid the HO penalty.

Fig. \ref{fig:velocities-rewards} depicts the average communication and transportation rewards and the HO rate for targeted speed $v=10$ m/s. It compares the performance of BDQN and BDDQN to those of the DDQN and the Shortest Distance Based-BS selection (SDB) benchmarks. From Fig. \ref{fig:velocities-rewards}, BDDQN and BDQN perform better than the benchmarks for all performance metrics. 
For instance, BDDQN outperforms SDN by $16.7\%$, $23.4\%$, and $10.9\%$, in terms of average transportation reward, average communication reward, and average HO rate, respectively.


Fig. \ref{fig:nrbss-rewards} illustrates the average communication reward, average transportation reward, and average HO rate for different numbers of BSs. The comparison is realized for BDQN, BDDQN, and SDN. As the number of available BSs along the 3D highway increases, the transportation reward decreases for BDQN and BDDQN, while it is slightly the same for SDB, as shown in Fig. \ref{fig:nrbss-rewards}(a). Indeed, the availability of more BSs encourages BDQN and BDDQN to increase further the communication reward and the expense of transportation reward, while SDN keeps the same strategy of connecting to the closest available BS all the time, as remarked in Fig. \ref{fig:nrbss-rewards}(b). 
For the proposed algorithms, an optimal number of BSs is identified. Specifically, with 15 BSs, the communication reward achieves its best values. However, increasing further the number of BSs is counter-productive as it causes more HO-related penalties.
Finally, 
Fig. \ref{fig:nrbss-rewards}(c) presents the HO rate for BDDQN when only 5 BSs or 15 BSs are available along the 3D highway. Surprisingly, when the number of BSs $N_R=5$ BSs, the HO rate is higher than that for $N_R=15$ BSs. Indeed, with a small BSs' number, the algorithm focuses on improving its transportation reward, e.g., reaching its target speed, thus causing the system to trigger HO more often.

\section{Conclusion}
In this work, we proposed BDQN and BDDQN algorithms to jointly optimize the network selection and autonomous moving actions such as changing the speed of the UAVs and switching the lanes to maximize both HO-aware data rate and 3D aerial highway traffic flow.
In the future, we will focus more on UAVs' precise positioning in intelligent transportation systems.
\bibliographystyle{IEEEtran}
\bibliography{main.bib}
\end{document}